\documentclass[letterpaper, 10 pt, conference]{ieeeconf}  

\IEEEoverridecommandlockouts                              

\usepackage{graphics} 
\usepackage{epsfig} 
\usepackage{times} 
\usepackage{amsmath} 
\usepackage{amssymb}  
\usepackage{cite}
\usepackage{url}
\usepackage{dblfloatfix}    
\usepackage[flushleft]{threeparttable}
\newcommand{\cmmnt}[1]{\ignorespaces}
\title{\LARGE \bf
Improved Planetary Rover Inertial Navigation and Wheel Odometry Performance through Periodic Use of Zero-Type Constraints
}

\author{Cagri Kilic, Jason N. Gross, Nicholas Ohi, Ryan Watson, Jared Strader, \\ Thomas Swiger, Scott Harper, and Yu Gu 
\thanks{This work was supported in part by the National Aeronautics and Space
Administration under Cooperative Agreement 80NSSC17M0053}
\thanks{Authors are with the Department of Mechanical and Aerospace Engineering, West Virginia University, Morgantown, WV, 26506. {\tt\small \{cakilic@mix.wvu.edu\}}} 
}

\begin{document}

\begin{onecolumn}
\Huge{\textbf{IEEE Copyright Notice}}

\vspace{5em}

\large{\textcopyright 2019 IEEE. Personal use of this material is permitted. Permission from IEEE must be obtained for all other uses, in any current or future media, including reprinting/republishing this material for advertising or promotional purposes, creating new collective works, for resale or redistribution to servers or lists, or reuse of any copyrighted component of this work in other works.}

\vspace{10em}

\Large{Accepted to be published in: Proceedings of the 2019 IEEE/RSJInternational Conference on Intelligent Robots and System (IROS 2019), November 4 - 8, 2019 in The Venetian Macao, Macau, China}

\end{onecolumn}

\twocolumn

\maketitle

\thispagestyle{empty}
\pagestyle{empty}

\begin{abstract}
We present an approach to enhance wheeled planetary rover dead-reckoning localization performance by leveraging the use of zero-type constraint equations in the navigation filter. Without external aiding, inertial navigation solutions inherently exhibit cubic error growth. Furthermore, for planetary rovers that are traversing diverse types of terrain, wheel odometry is often unreliable for use in localization, due to wheel slippage. For current Mars rovers, computer vision-based approaches are generally used whenever there is a high possibility of positioning error; however, these strategies require additional computational power, energy resources, adequate features in the environment, and significantly slow down the rover traverse speed. To this end, we propose a navigation approach that compensates for the high likelihood of odometry errors by providing a reliable navigation solution that leverages non-holonomic vehicle constraints as well as state-aware pseudo-measurements (e.g., zero velocity and zero angular rate) updates during periodic stops. By using this, computationally expensive visual-based corrections could be performed less often. Experimental tests that compare against GPS-based localization are used to demonstrate the accuracy of the proposed approach. The source code, post-processing scripts, and example datasets associated with the paper are published in a public repository. 

\end{abstract}

\section{INTRODUCTION}
The Mars Sample Return (MSR) mission was identified as the highest priority planetary science objective in the Planetary Science Decadal Survey \cite{board2012vision}. To meet the requirement of the sample return missions, a mobile rover should be able to traverse the terrain safely, quickly, and autonomously to reach its desired targets. In particular, highly accurate real-time localization performance is one of the most demanding abilities for autonomous driving \cite{c1}, which is especially challenging for planetary rovers with limited energy resources and computational power due to radiation-hardened hardware requirements. 

Previous Mars rovers have exhibited large onboard localization errors and have had to rely primarily on human-in-the-loop operations \cite{c1}. The Mars Exploration Rovers (MERs) and Mars Science Laboratory (MSL) use a similar approach for localization. When moving, the rover's attitude angles are propagated through the integration of angular-rate measurements \cite{c4}. When the rover is stopped, the rover's pitch and roll angles are estimated based on the gravity vector which is measured using tri-axial accelerometers \cite{c4}. Rover position is estimated from wheel encoder-based odometry and inertial measurement unit (IMU) integration \cite{c5}. Stereo vision-based visual odometry (VO) \cite{c6} is also used whenever a high probability of wheel slippage is anticipated. Due to limited onboard computational resources, to refine rover position estimates, incremental bundle adjustment is performed on Earth by post-processing downlinked images \cite{c7}.

Wheel slippage is one of the most critical issues to be dealt with for mobile robots driving across loose soil \cite{c12}. For example, MERs both became embedded into the soft surface of Mars \cite{c18,c13}, due to the significant amount of slip. In May 2009, Spirit became permanently entrapped in soft soil \cite{c15}. Moreover, Curiosity had to stop to avoid sinking because of severe wheel slippage \cite{c17}. 

One of the most common techniques for estimating rover slip is based on VO \cite{c12} \cite{c64}. Although VO is an accurate source of information for slip estimation, it is computationally expensive. For instance, each step required 2-3 min of computation time on the MER's 20-MHz CPU \cite{c5}. Although MSL's 200-MHz CPU has reduced the onboard processing time to under 40 seconds per step \cite{c62} \cite{c54}, the other limitation of VO arises on low-feature terrains (e.g., sand dunes, shadowed areas). A low number of detected and tracked features can lead to poor accuracy of motion estimate \cite{c21}.  

For the proposed future MSR mission \cite{c66}, the rover may be required to traverse up to 20 km from the lander to the cached samples (left by the Mars 2020 rover) and back to its landing area in less than 99 sols. This required traversal rate is much faster than Curiosity, which has traveled 19.75 km in 6 years \cite{c67} and Opportunity that had traversed 45.16 km in 15 years \cite{c68} on Mars before the Planet-Encircling Dust Event. It is possible that the MSR rover will need to travel up to 1 km per day to the cached sample, which is well beyond the capability for the current daily operations planning. This increased rover autonomy requirement, in turn, requires more accurate and computationally efficient onboard rover localization capabilities.  

Zero-type updates (i.e., zero velocity and zero angular rate) are widely used to aid inertial pedestrian navigation \cite{foxlin2005, norrdine2016}. Observability characteristics of zero-type updates on a land vehicle are also detailed in \cite{ramanandan2011}. Zero velocity detection and application in standard road conditions is shown in \cite{xiaofang2014}. Similarly, the method in \cite{ramo} demonstrates decreased growth in state errors for a land vehicle when stationary updates are used; e.g., stopping at traffic lights.
\begin{figure*}[b!]
\centering
\includegraphics[scale=0.4]{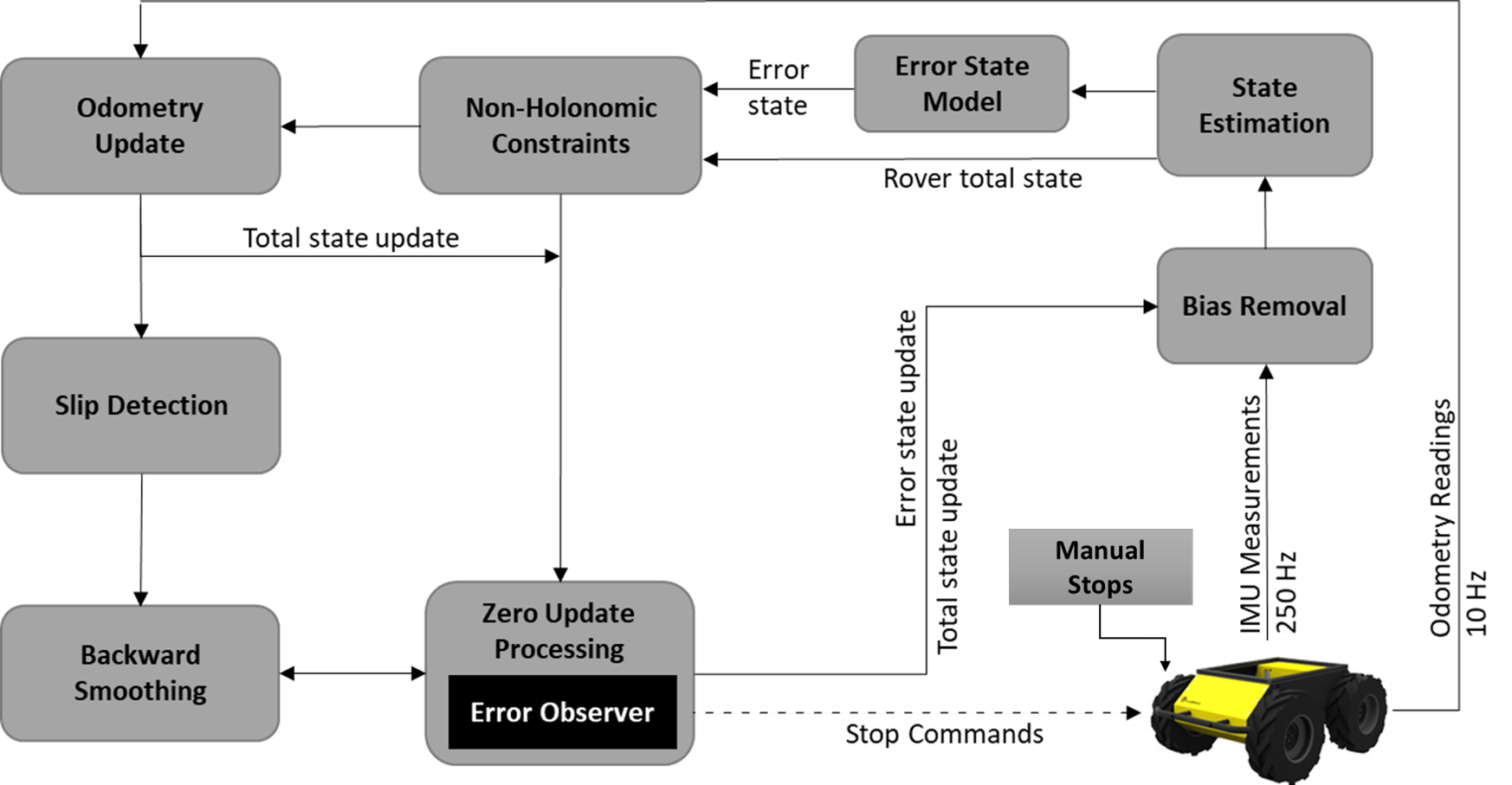}
\caption{This figure presents a graphical depiction of the proposed navigation system when all the algorithm components are utilized. Within the proposed algorithm, the IMU observations are used to propagate the current state estimate. Wheel odometry observations, in conjunction with non-holonomic constraints, are leveraged to reduce the unbounded error growth of the un-aided inertial navigation system. The zero-type updates can be triggered by an error observer or manual stops.}
\label{figurelabel2}
\end{figure*}
NASA's Mars rovers move slowly (e.g., MSL has 0.04 m/s top speed on flat hard ground \cite{rover1}), stop often, and are also equipped with high quality LN-200S IMUs \cite{bies}. These facts point to the potential that rover localization performance stands to be improved by leveraging zero-type updates for the Curiosity rover as well as the future Mars 2020 and MSR rovers without any hardware changes or significant changes in rover operations. 

 \section{METHODOLOGY}
The architecture of the proposed approach is given in Fig.~\ref{figurelabel2}. The proposed approach offers the following contributions: 1) it is demonstrated that zero-type updates, which simply require stopping, can significantly reduce the rate of inertial navigation error growth for wheeled rovers, 2) we show that wheel slip can be detected as velocity discrepancies between wheel odometry measurements and inertial navigation system (INS) solution, and 3) to show the benefit of the proposed approach, we present experimental tests in which dead-reckoning localization estimates are compared to a differential GPS (DGPS) reference solution. As a final contribution, we also make our software, which is designed for use under ROS \cite{ROS}, and presented datasets publicly available.

To improve inertial navigation and provide uncertainty bounds, an error-state extended Kalman filter (EKF) is implemented based on the method detailed in \cite{grovebook}. In this formulation, a planet-centered, planet-fixed frame is used as the reference frame, while a locally-level-navigation frame (NED) comprises the resolving axes. In the filter, wheel odometry based wheel velocities and heading rate calculations are treated as an aiding sensor for the INS. In addition, zero-type updates and non-holonomic motion constraints are utilized as additional pseudo-measurement updates whenever they are available. Slip is estimated by using a threshold on the difference between the estimated INS solution and estimated wheel odometry velocity vectors. In this paper, navigation stops are triggered by a manual stopping frequency. Finally, during each navigation stop, backward smoothing to the previous navigation stop is used to refine state estimates further.
 \subsection{State Updates}
The navigation filter implementation is composed of an attitude update, a velocity update, and a position update. We briefly summarize the integration equations here for completeness. Detailed descriptions can be found in \cite{grovebook}. The attitude update is given as
\begin{equation}
\label{eq:su2}
\begin{aligned}
{\mathbf{C}_{b}^{n}}^{(+)}\approx{} & {\mathbf{{C}}_{b}^{n}}^{(-)}  \bigl (\mathbf{I}_3 + \mathbf{\Omega}_{ib}^{b} \Delta t_i \bigr ) \\
& -\bigl ({\mathbf{\Omega}_{ie}^{n}}^{(-)} +{\mathbf{\Omega}_{en}^{n}}^{(-)}  \bigr )\mathbf{{C}}_{b}^{n} \Delta t_i
\end{aligned}
\end{equation}
where  $\mathbf{C}_{b}^{n}$ is the coordinate transformation matrix from the body frame to the locally level frame, $\mathbf{I_3}$ is a 3-by-3 identity matrix, $\mathbf{\Omega}_{ib}^{b}$ is the skew symmetric matrix of the IMU angular rate measurement, $\mathbf{\Omega}_{ie}^{n}$ is the skew symmetric matrix of the planet's rotation vector represented in the locally level frame, $\mathbf{\Omega}_{en}^{n}$ is the transport term, and $\Delta t_i$ is the IMU sampling interval.

Assuming that the variations of the acceleration due to gravity, Coriolis, and transport rate terms are all negligible over the integration interval, the velocity update is given as,
\begin{equation}
\begin{aligned}
{\mathbf{v}_{eb}^{n}}^{(+)} \approx {} & {\mathbf{v}_{eb}^{n}}^{(-)} +(\mathbf{f}_{ib}^{n}+\mathbf{g}_{b}^{n}(L_b^{(-)} ,h_b^{(-)} ) \\
& - ( {\mathbf{\Omega}_{en}^{n}}^{(-)} +2{\mathbf{\Omega}_{ie}^{n}}^{(-)} ){\mathbf{v}_{eb}^{n}}^{(-)} )\Delta t_i 
\end{aligned}
\end{equation}
where $\mathbf{v}_{eb}^{n}$ is the velocity update, $\mathbf{f}_{ib}^{n}$ is the specific force measurements from the IMU acceleration sensors, $\mathbf{g}_{b}^{n}$ is the gravity vector.

Finally, assuming the velocity variation is linear over the integration interval, the position update is given as
\begin{equation}
h_b^{(+)} =h_b^{(-)} -\frac{\Delta t_i}{2} \biggl({\mathbf{v}_{eb,D}^{n}}^{(-)}  +{\mathbf{v}_{eb,D}^{n}}^{(+)}  \biggr)
\end{equation}

\begin{equation}
\begin{aligned}
L_b^{(+)} =L_b^{(-)} +\frac{\Delta t_i}{2} \frac{{\mathbf{v}_{eb,N}^{n}}^{(-)} }{R_N(L_b^{(-)} )+h_b^{(-)} } \\+\frac{\Delta t_i}{2}\frac{{\mathbf{v}_{eb,N}^{n}}^{(+)} }{R_N(L_b^{(-)} )+h_b^{(+)} } 
\end{aligned}
\end{equation}

\begin{equation}
\begin{aligned}
\lambda_b^{(+)} =\lambda_b^{(-)} +\frac{\Delta t_i}{2} \frac{{\mathbf{v}_{eb,E}^{n}}^{(-)} }{\bigl(R_E(L_b^{(-)} )+h_b^{(-)} \bigr)cosL_b^{(-)} } \\ +\frac{\Delta t_i}{2}\frac{{\mathbf{v}_{eb,E}^{n}}^{(+)}}{\bigl(R_P(L_b^{(-)})+h_b^{(+)}\bigr)cosL_b^{(+)}}
\end{aligned}
\end{equation}
where $h_b$, $L_b$ and $\lambda_b$ are updated position estimates (expressed in terms of height, latitude, and longitude, respectively), $R_N$ is the variation of the meridian, and $R_P$ is transverse radii of curvature.

\subsection{Error State Model}
To calculate the INS system matrix, the Jacobian of the error-state equations is determined. After defining the time derivatives of the error-state equations, the system matrix and the state transition matrix are defined. Then, the errors are transformed into the local navigation frame, and the time derivative of the velocity error is constructed by adding the transport rate term. The error state vector is constructed in a local navigation frame, 
\begin{equation}
\label{errorstate}
\mathbf{x}_{err}^{n}={\biggl(
\delta\mathbf{ \Psi}_{nb}^{n} \ \ 
\mathbf{\delta v}_{eb}^{n} \ \ 
\delta\mathbf{p}_{b} \ \ 
\mathbf{b}_a \ \ 
\mathbf{b}_g
\biggr )}^{\mathbf{T}}
\end{equation}
where, $\delta\mathbf{ \Psi}_{nb}^{n}$ is the attitude error, $\mathbf{\delta v}_{eb}^{n}$ is the velocity error, $\delta\mathbf{p}_{b}$ is the position error, $\mathbf{b}_a$ is the IMU acceleration bias, and $\mathbf{b}_g$ is the IMU gyroscope bias.
The position error is expressed in terms of the latitude, longitude, and height, respectively. 
\begin{equation}
\quad 
\delta \mathbf{p}_{b}={\biggl(
\delta \mathbf{L}_{b} \ \ 
\delta \mathbf{\lambda}_{b} \ \
\delta \mathbf{h}_{b}
\biggr)}^{\mathbf{T}}
\end{equation}
The interested reader is referred to \cite{grovebook} for a detailed derivation of the INS error-state model adopted in this work. 
\subsection{Non-Holonomic Constraints}
\label{nonholo}
A rover is a non-holonomic system if its number of controllable degrees of freedom is less than its total degrees of freedom. A skid-steer rover, such as a Clearpath Robotics Husky \cite{Husky}, is subject to two motion constraints if the rover is not experiencing side slip and motion normal to the road surface: 1) the velocity of the vehicle is zero along the rotation axis of any of its wheels, and 2) velocity is also zero in the direction perpendicular to the traversal surface \cite{diss}. Due to frame rotations, zero vertical and lateral velocity does not mean that acceleration on these directions are zero.  

Following the similar process as in \cite{grovebook}, it is assumed that the error-state vector is defined by \eqref{errorstate} and the total state vector is 
\begin{equation}
\mathbf{x}^{n}={\biggl(
\mathbf{\Psi}_{nb}^{n} \ \ 
\mathbf{v}_{eb}^{n} \ \ 
\mathbf{p}_{b} 
\biggr )}^{\mathbf{T}} .
\end{equation}
The rover velocity constraints can be applied as a pseudo-measurement update, assuming that the axes of the rear-wheel frame are aligned with the body frame. This measurement update may be expressed as
\begin{equation}
\mathbf{\delta z}_{RC}^{n}=-\begin{pmatrix}
0 & 1 & 0\\ 
0 & 0 & 1
\end{pmatrix} (\mathbf{C}_{n}^{b} \mathbf{v}_{eb}^{n} -\mathbf{\omega}_{ib}^{b} \times \mathbf{L}_{rb}^{b})
\end{equation}
where $\mathbf{L}_{br}^{b}$ is body to rear wheel lever arm, and $\mathbf{\omega}_{ib}^{b}$ is angular rate measurement. Then, corresponding measurement matrix may be approximated as
\begin{equation}
\mathbf{H}_{RC}^{n}=\begin{pmatrix}
\mathbf{0}_{2,3} & \begin{bmatrix}-\mathbf{H}_{l} \\ -\mathbf{H}_{v}\end{bmatrix} &\mathbf{0}_{2,3}  &\mathbf{0}_{2,3} &\mathbf{0}_{2,3}
\end{pmatrix}
\end{equation}
where $\mathbf{H}_{l}$ is lateral constraint part. and $\mathbf{H}_{v}$ is the vertical part of the measurement matrix. 
\begin{equation}
\begin{bmatrix}-\mathbf{H}_{l} \\ -\mathbf{H}_{v}\end{bmatrix} = \begin{pmatrix}
0 & 1 & 0 \\ 0 & 0 & 1 \end{pmatrix} \mathbf{C}_{n}^{b}
\end{equation}

Although the non-holonomic constraint measurement update is performed whenever the navigation solution is updated, note that the excessive sideslip invalidates the lateral velocity constraint, and this adds extra biases to velocity solution. While turning, the heading rate of the rover can be observed, and if it exceeds a threshold, the lateral velocity constraint measurement can be omitted.
\subsection{Zero Type Updates} 
\label{zeroUpd}
During stationary conditions, IMU sensor outputs are governed by the planet's rotation and sensor errors. Since the rover is often stationary during the mission, zero-type updates can be leveraged to maintain the INS alignment. To properly use these updates, stationary conditions must be detected accurately. Otherwise, the rover's state yields incorrect updates leading to poor navigation performance \cite{ramo}. To detect stationary conditions, we used two different indicators, wheel odometry velocity, and the standard deviation of the IMU measurements, which assumes that the rover typically experiences less vibration whenever it is stationary. 

Zero-type updates bound the velocity error and calibrate IMU sensor noises \cite{skog}. Therefore, the measurement innovation for a zero-type update can be given as 
 \begin{equation}
 \mathbf{\delta z}_{Z,k}^{n -}=[-\mathbf{\hat{v}}_{eb,k}^{n},-\mathbf{\hat{\omega}}_{ib,k}^{b}]^{T} 
 \end{equation}
where $\mathbf{\delta z}_{Z,k}^{n -}$ is measurement innovation matrix, $\mathbf{\hat{v}}_{eb,k}^{n} $ is estimated velocity vector, and $\mathbf{\hat{\omega}}_{ib,k}^{b}$ estimated gyro bias. The measurement matrix is given
\begin{equation}
\mathbf{H}_{Z,k}^{n}=\begin{bmatrix}  \mathbf{0}_3 & \mathbf{-I}_3 & \mathbf{0}_3 & \mathbf{0}_3 & \mathbf{0}_3\\
\mathbf{0}_3 & \mathbf{0}_3 & \mathbf{0}_3 & \mathbf{0}_3 & \mathbf{-I}_3
\end{bmatrix}\end{equation}

It is also important to note that zero-type updates do not provide any position information; however, because the error-state model properly accounts for the correlation between the velocity and position errors in the off-diagonal elements of the error covariance matrix, the cubic error growth of INS positioning is reduced to linear growth \cite{mather2006man}. This enables zero-type updates to not only limit the error growth and help determine biases to reduce future error growth, but also to mitigate most of the position drift since the last navigation stop. 

\subsection{Wheel Odometry Update}
The wheel odometry outputs, wheel forward speed, and heading rate are averaged over the odometry measurement sampling interval. Heading rate is calculated by differencing the average linear velocity values of left and right wheels. It is also assumed that the axes of the rear and front wheel frames are aligned with the body frame. 

In our implementation, we adopt a similar procedure as described in \cite{diss} and \cite{grovebook} to aid the sensors with an additional vertical velocity constraint as  
\begin{equation}
\delta \mathbf{z}_{O}=\begin{pmatrix}
{\tilde{v}}_{lon,O}- {\tilde{v}}_{lon,i}\\ -{\tilde{v}}_{lat,i}\\-{\tilde{v}}_{ver,i}\\ 
\tilde{\dot{{\psi}}}_{nb,o}-\tilde{\dot{{\psi}}}_{nb,i}\overline{cos{\hat{\theta}_{nb}}}
\end{pmatrix}
\end{equation}
\begin{equation}
\begin{aligned}
\begin{bmatrix}
{\tilde{v}}_{lat,i}\\ {\tilde{v}}_{lon,i}\\ {\tilde{v}}_{ver,i}
\end{bmatrix}=&\frac{1}{\tau_0}\int_{t-\tau_{0}}^{t} \mathbf{I}_{3}\big(\mathbf{C}_{n}^{b}(\tau) \mathbf{v}_{eb}^{n}(\tau) \\
& +\mathbf{w}_{eb}^{b}(\tau)  \times \mathbf{L}_{br}^{b}\big)d\tau
\end{aligned}
\end{equation}
where $\tilde{v}_{lon}$, $\tilde{v}_{lat}$, and $\tilde{v}_{ver}$ are predicted longitudinal, lateral and vertical rear wheel speed, respectively. The subscript $i$ denotes the estimated INS solution, and $O$ denotes the wheel odometry measurements. The measurement matrix is
\begin{equation}
 \mathbf{H}_{O,k}^{n} = \begin{bmatrix} 
 \begin{bmatrix}\mathbf{H}_{O1:3,1}^{n}\end{bmatrix}& [\mathbf{H}_{O1:3,2}^{n}] &   &\mathbf{0}_{9}   &\ \\
 \mathbf{H}_{O4,1}^{n}&\mathbf{0}_3  &\mathbf{0}_3  &\mathbf{H}_{O4,4}^{n}  &\mathbf{0}_3
\end{bmatrix} 
\end{equation}
where
\begin{equation}
\begin{aligned}
 \mathbf{H}_{O1:3,1}^{n} =& -\frac{1}{\tau_0}\int_{t-\tau_{0}}^{t} \mathbf{I}_{3}\mathbf{C}_{n}^{b}(\tau) [\mathbf{v}_{eb}^{n}(\tau)  \times ]d\tau,  \\
 \mathbf{H}_{O1:3,2}^{n} =& -\frac{1}{\tau_0}\int_{t-\tau_{0}}^{t} \mathbf{I}_{3}\mathbf{C}_{n}^{b}(\tau) d\tau,  \\
 \\
  \mathbf{H}_{O4,1}^{n} =& \frac{1}{\tau_0^{2}}[\hat{\psi}_{nb}(t)-\hat{\psi}_{nb}(t-\tau_{0})]\\
&\int_{t-\tau_{0}}^{t} sin\hat{\theta}_{nb}\begin{bmatrix}
0\\ cos\hat{\phi}_{nb}\\ sin\hat{\phi}_{nb}
\end{bmatrix}^{T}\mathbf{C}_{n}^{b}(\tau)d\tau,
\\
\mathbf{H}_{O4,4}^{n} = &- \frac{cos{\hat{\theta}_{nb}}}{\tau_0}\begin{bmatrix}
0\\ 0\\ 1
\end{bmatrix}^{T} \mathbf{C}_{b}^{n}
\end{aligned}
\end{equation}
where $\mathbf{0}_{9}$ is a 9-by-9 zero matrix, $\mathbf{H}_{O1:3,1}^{n}$ and $\mathbf{H}_{O1:3,1}^{n}$ are 3-by-3 matrices,  $\mathbf{H}_{O4,1}$ and $\mathbf{H}_{O4,4}$ are 1-by-3 row vectors. Note that the coordinate transformation matrix in $\mathbf{H}_{O44}^{n}$ is from the body to the inertial navigation frame. 


\subsection{Slip Detection}
\label{slipdetection}
Skid steer vehicles rely on differing left and right wheel velocity directions to turn the vehicle. Due to redundant points of contact (i.e., two wheels are driven by the same drive-train on each side), slippage is often expected when turning motion is performed \cite{ebersolephd}. 
Since the IMU measurements are independent of the wheel odometry, the motion estimates from the EKF can be compared to the computed velocity based on the vehicle kinematics to determine if any statistically significant slippage has occurred. In order to detect the anomalies in velocity, the residuals between wheel odometry measurements and the EKF solution are computed using the Mahalanobis distance and compared against a threshold. Post-fit residuals from the wheel odometry update are used to calculate the Mahalanobis distance as given in (\ref{mahalanobis}).

\begin{equation}
\label{mahalanobis}
\chi=\sqrt{(\Tilde{\mathbf{z}}-\Tilde{\mathbf{H}}\mathbf{x}_{err})^{T}\mathbf{S}^{-1}(\Tilde{\mathbf{z}}-\Tilde{\mathbf{H}}\mathbf{x}_{err})}
\end{equation}
where $\mathbf{S}$ is the predicted covariance matrix. An empirical threshold on the Mahalanobis distance, $\chi$, is used to specify whether the velocity anomaly has occurred or not.

The slippage is also monitored with the slip ratio calculation for front and rear wheels with respect to the INS velocity solution. The slip ratio, $i \in[-1,1]$, is checked to specify if the slip is significant (i.e., $|i|>0.3$) along with the Mahalanobis distance value at the same time-step. We define the slip ratio, i, as follows:
\begin{equation}
\label{slip}
i= 1-\frac{v_x}{r\omega}
\end{equation}
where ${v_x}$ is the translational velocity estimated from INS, $r$ is the wheel radius, and $\omega$ is the wheel angular velocity estimated from the wheel odometry measurements.  
\subsection{Backward Smoothing}
To smooth out the estimated state along the traversed path between zero-type updates, we used the Rauch-Tung-Striebel (RTS) backward recursive method \cite{c76}, which utilizes all the available state information and goes backward in time.

A backward smoothing algorithm can be summarized as
\begin{equation}
\mathbf{A}_{k}=\mathbf{P}_{(k,k)}\mathbf{\Phi}_{k}^{T}(\mathbf{P}_{(k+1,k)})^{-1}
\end{equation}
\begin{equation}
\mathbf{x}_{k,s}=\mathbf{x}_{(k,k)}\mathbf{A}_{k}(\mathbf{x}_{(k+1),s}-\mathbf{x}_{(k+1,k)})
\end{equation}
\begin{equation}
\mathbf{P}_{k,s}=\mathbf{P}_{(k,k)}\mathbf{A}_{k}(\mathbf{P}_{(k+1),s}-\mathbf{P}_{(k+1,k)})
\end{equation}
where $\mathbf{A}$ is the smoothing gain, $\mathbf{x}_{k,s}$ and $\mathbf{P}_{k,s}$ are smoothed state vector and error covariance, respectively. Although zero-type updates correct most of the drift when the rover is stopped, the rover's position solution still drifts while driving. To utilize the backward smoothing algorithm, the estimated states, state transition matrix, and error covariance between zero-type updates are stored in memory. The smoothing algorithm starts from the current navigation stop and goes back to the most recent previous navigation stop. This further mitigates the solution drift between stopping. 

After smoothing between stops, the forward filter continues to provide a real-time available state estimate until the next navigation stop.  In this scenario, for error analysis, it is not simple to present the solution that would be available for real-time over a complete trajectory that includes multiple navigational stops. Therefore, in the presented experimental evaluation, solutions that only include the forward filtering mode are shown, so that the impact of smoothing can be assessed.

\section{EXPERIMENTAL RESULTS}
For experimental evaluation, a reference solution was determined using carrier-phase DGPS. This set-up consisted of two dual-frequency Novatel OEM-615 GPS receivers \cite{novatel1} and dual-frequency antennas, with one set mounted on a static base station and another affixed on top of the test rover platforms. During the experiments, 10 Hz raw GPS pseudorange and carrier-phase measurements were recorded on both receivers, and the reference position solutions were then post-processed using the open-source software library, RTKLIB 2.4.2 \cite{rtklib}. Post-processed carrier-phase DGPS is expected to provide centimeter-to-decimeter level accuracy \cite{gps}.

The IMU incorporated on the rovers is an ADIS 16495-2 \cite{adis}. This IMU reports to have an in-run bias, and angular random walk values are $1.6^{o}/hr$, $0.1^{o}/\sqrt{hr}$ for the gyroscopes, and a bias-stability and velocity random walk of 3.2$\mu g$ and 0.008 m/sec/$\sqrt{hr}$ for the accelerometers, respectively. For wheel odometry readings, the Clearpath Robotics Husky's quadrature encoders with 78,000 pulses/m, and WVU Pathfinder Test Platform's quadrature encoders with 47,000 pulses/m resolution are used. 

\begin{figure}[thbp]
      \centering
      \includegraphics[width=\columnwidth]{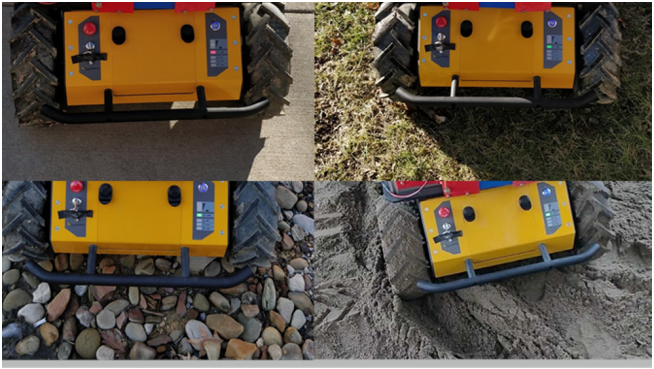}
      \caption{Husky Rover on traversed terrain types to test the developed navigation approach.}
      \label{terrains}
      \end{figure}

\begin{figure}[thbp]
      \centering
      \includegraphics[width=\columnwidth]{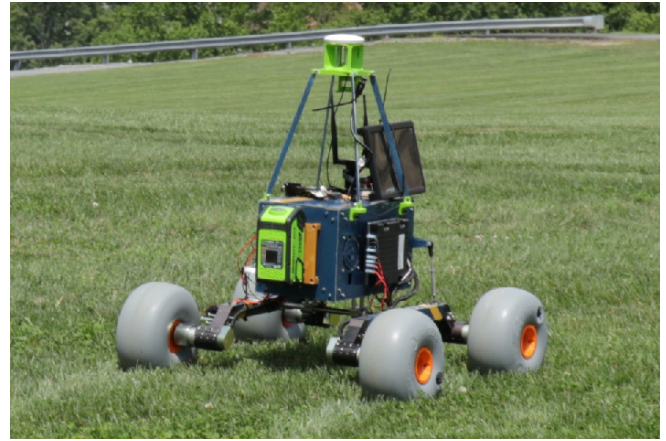}
      \caption{WVU Pathfinder Test Platform on non-flat lawn.}
      \label{pf}
      \end{figure}

Several tests on different terrain types, i.e., concrete, grass, gravel, and sand, as shown in Fig.~\ref{terrains}, were performed on the West Virginia University campus using two different rovers. These rovers, Clearpath Husky A200 and WVU Pathfinder Test Platform are shown in Figs.~\ref{terrains}-\ref{pf}, respectively.
      
To demonstrate the proposed navigation approach, four different scenarios are detailed: 
1) {\textbf{``Concrete-Turn"}}, the Husky rover follows a flat, L-Shaped concrete path.
2) {\textbf{``Rough-Terrain"}}, the Husky rover traverses a longer path on non-flat, muddy, and grassy terrain with embedded rocks.
3) {\textbf{``Fast-Rectangle"}}, the Pathfinder rover traverses a rectangle path on non-flat, grassy terrain with high speed.
4) {\textbf{``Slow-Rectangle"}}, the Pathfinder rover traverses a rectangle path on non-flat, grassy terrain with slow speed.
For each scenario, navigation stops were periodically commanded. The total distance traversed, Average Driving Time between Stops (ADTS), the number of performed stops, and the total driving time are listed in Table \ref{tableScene}. The Husky rover has 0.4 m/s commanded velocity while driving in Concrete-Turn and Rough-Terrain scenarios. The Pathfinder rover has 0.8 m/s, and 0.2 m/s average commanded velocity in Fast-Rectangle and Slow-Rectangle scenarios, respectively.
 

To assess the relative benefit of each aspect of the approach, many of the update-type solution combinations are performed for each of the test runs. The positioning error values with respect to each of the update combinations are given for each scenario in Tables \ref{tab:lshape} - \ref{tab:fastrun}, where RMS denotes root mean square, STD denotes standard deviation of the horizontal error, and Max. is the worst-case horizontal distance error that rover encounters during each test.

\begin{table}
\footnotesize
\begin{threeparttable}
\caption{Details of Performed Scenarios}
\label{tableScene}
\centering
\begin{tabular}{@{}lcccc@{}}
\hline
Scenario Type &$\Sigma$ Distance &ADTS & $\Sigma$ Stops &$\Sigma$ Driving Time \\
\hline\hline
Concrete-Turn  &34m   &12s   &7    &85s  \\
Rough-Terrain  &151m  &9s	 &42   &406s \\
Fast-Rectangle  &87m  &15s	 &8   &120s \\
Slow-Rectangle  &85m  &15s	 &21   &330s \\
\hline
\end{tabular}
\end{threeparttable}
\end{table} 

\begin{table} [h!]
\footnotesize
\begin{threeparttable}
\caption{Position Error for Concrete-Turn Scenario}
\label{tab:lshape}
\centering
\begin{tabular}{@{}lcccccc@{}}
\hline
Estimation & \multicolumn{3}{c}{Horizontal Error (m)}   &\multicolumn{3}{c}{RMS Error (m)}\\
 I O Z N B \tnote{*}& \scriptsize{Median}& \scriptsize{STD}& \scriptsize{Max.}& \scriptsize{East}& \scriptsize{North}& \scriptsize{Up}\\ 
\hline\hline
I               &518.44     &494.77   &1606.2   &783.75     &64.75	    &153.07 \\
I+O             &23.73	    &12.59	  &34.26	&21.80	    &10.08	    &7.07\\
I+O+N           &23.69	    &11.89	  &32.34	&20.57      &10.57	    &7.84\\
I+O+N+B         &23.75	    &12.25	  &33.30	&21.18	    &10.45	    &7.57\\
I+Z             &2.86	    &6.42	  &31.09	&7.38	    &4.28	    &4.46\\
I+Z+B           &0.60	    &1.58	  &31.62	&1.23	    &1.46	    &1.68\\
I+Z+N           &0.58	    &1.19	  &8.54	    &1.28	    &0.81       &0.81\\
I+Z+N+B         &0.63	    &1.13	  &7.69	    &1.28	    &0.74	    &0.95\\
I+Z+O           &0.68	    &0.31	  &1.33	    &0.67	    &0.31	    &0.67\\
I+Z+B+O        &0.66	    &0.31	  &1.33	    &0.66	    &0.31	    &0.69\\
I+Z+N+O        &0.49	    &0.25	  &1.20	    &0.43       &0.36	    &0.71\\
I+Z+N+B+O       &0.48	    &0.25	  &1.21	    &0.43	    &0.36	    &0.73\\
\hline
\end{tabular}
\begin{tablenotes}
\item[*] I:INS, O:Odometry, Z:Zero-Type, N:Non-Holonomic, B:Backward \\Smoothing.
\end{tablenotes}
\end{threeparttable}
\end{table}

\begin{table}[h!]
\footnotesize
\begin{threeparttable}
\caption{Position Error for Rough-Terrain Scenario}
\label{tab:longrun}
\centering
\begin{tabular}{@{}lcccccc@{}}
\hline
Estimation & \multicolumn{3}{c}{Horizontal Error (m)}   &\multicolumn{3}{c}{RMS Error (m)} \\
I O Z N B \tnote{*}& \scriptsize{Median}& \scriptsize{STD}& \scriptsize{Max.}& \scriptsize{East}& \scriptsize{North}& \scriptsize{Up}\\
\hline\hline
I               &14253.	    &20742.	    &68676.	    &29200.	&6045.	&11334.\\
I+O             &21.31	    &18.69	    &63.90	    &23.37	&17.45	&18.07\\
I+O+N           &30.68	    &16.30	    &56.53	    &29.23	&16.94	&11.65\\
I+O+N+B         &30.84	    &16.35	    &55.71	    &28.92	&17.06	&11.97\\
I+Z             &1.77	    &3.13	    &31.40	    &4.02	&1.09	&2.19\\
I+Z+B           &1.55	    &1.23	    &29.75	    &2.06	&0.96	&0.86\\
I+Z+N           &0.93	    &0.99	    &4.08	    &1.46	&0.73	&0.92\\
I+Z+N+B         &1.59	    &1.07	    &4.01	    &1.79	&1.09	&0.58\\
I+Z+O           &1.36	    &1.29	    &4.28	    &2.16	&0.29	&1.18\\
I+Z+B+O         &1.21	    &1.28	    &4.29       &2.13	&0.32	&0.90\\
I+Z+N+O         &0.47	    &0.90	    &2.86	    &1.25	&0.27	&1.04\\
I+Z+N+B+O       &0.54	    &0.93   	&2.89	    &1.28	&0.41	&0.80\\
\hline
\end{tabular}
\begin{tablenotes}
\item[*] The same estimation types are used as in Table \ref{tab:lshape}.
\end{tablenotes}
\end{threeparttable}
\end{table}
    
    \begin{table}[h!]
\footnotesize
\begin{threeparttable}
\caption{Position Error for Fast-Rectangle Scenario}
\label{tab:slowrun}
\centering
\begin{tabular}{@{}lcccccc@{}}
\hline
Estimation & \multicolumn{3}{c}{Horizontal Error (m)}   &\multicolumn{3}{c}{RMS Error (m)} \\
I O Z N B \tnote{*}& \scriptsize{Median}& \scriptsize{STD}& \scriptsize{Max.}& \scriptsize{East}& \scriptsize{North}& \scriptsize{Up}\\
\hline\hline
I               &7801.5    &8945.6	    &29937.	    &13482.	&1450.	&3783.\\
I+O             &15.85	    &8.83	    &30.43	    &13.97	&8.22	&8.52\\
I+O+N           &7.04	    &9.26	    &29.08	    &12.15	&6.08	&5.06\\
I+O+N+B         &4.90	    &7.00	    &22.19	    &9.38	&4.64	&6.71\\
I+Z             &5.42	    &10.27	    &73.16	    &12.18	&3.01	&5.04\\
I+Z+B           &4.20	    &4.85	    &118.28     &4.73	&4.52	&5.24\\
I+Z+N           &2.26	    &1.38	    &4.69	    &1.56	&1.82	&1.82\\
I+Z+N+B         &0.96	    &1.45	    &8.29	    &1.54	&1.35	&1.81\\
I+Z+O           &1.29	    &1.05	    &3.44	    &1.06	&1.33	&1.31\\
I+Z+B+O         &1.78       &1.06       &3.64       &1.47   &1.19   &0.62\\
I+Z+N+O         &1.34       &1.03       &3.50       &1.09   &1.33   &1.98\\
I+Z+N+B+O       &0.92	    &0.84   	&2.77	    &0.80	&1.06	&1.19\\
\hline
\end{tabular}
\begin{tablenotes}
\item[*] The same estimation types are used as in Table \ref{tab:lshape}.
\end{tablenotes}
\end{threeparttable}
\end{table}
    
    \begin{table}[h!]
\footnotesize
\begin{threeparttable}
\caption{Position Error for Slow-Rectangle Scenario}
\label{tab:fastrun}
\centering
\begin{tabular}{@{}lcccccc@{}}
\hline
Estimation & \multicolumn{3}{c}{Horizontal Error (m)}   &\multicolumn{3}{c}{RMS Error (m)} \\
I O Z N B \tnote{*}& \scriptsize{Median}& \scriptsize{STD}& \scriptsize{Max.}& \scriptsize{East}& \scriptsize{North}& \scriptsize{Up}\\
\hline\hline
I               &25771.	    &31574.	    &98873.	&46708.	&6222.	&14016.\\
I+O             &8.71	    &8.51	    &28.19      &14.22	&4.77	&4.09\\
I+O+N           &12.81	    &10.81	    &34.76	    &16.71	&9.28	&8.84\\
I+O+N+B         &10.95      &8.19	    &29.09	    &14.04	&3.48	&9.44\\
I+Z             &4.34	    &5.50	    &43.70	    &6.84	&3.81	&3.14\\
I+Z+B           &4.68       &2.66	    &97.09	    &3.80	&3.73	&2.93\\
I+Z+N           &3.17	    &1.23	    &6.89	    &1.73	&2.55	&1.08\\
I+Z+N+B         &1.77	    &1.24	    &13.44	    &1.95	&1.19	&1.04\\
I+Z+O           &3.74	    &1.13	    &4.61	    &2.09	&2.75	&1.88\\
I+Z+B+O         &3.99	    &1.19	    &5.05       &3.04	&2.40	&1.18\\
I+Z+N+O         &1.67	    &0.68	    &2.87	    &0.93	&1.51	&1.69\\
I+Z+N+B+O       &1.34	    &0.69   	&2.90	    &0.95	&1.34	&1.77\\
\hline
\end{tabular}
\begin{tablenotes}
\item[*] The same estimation types are used as in Table \ref{tab:lshape}.
\end{tablenotes}
\end{threeparttable}
\end{table}
Without using zero-type updates, the rover navigation performance is inadequate due to the accumulated error of the INS integration and the wheel slippage. Wheel odometry helps to reduce drift along with non-holonomic constraints within the INS solution; however, if the zero-type updates are applied, most of the accumulated error is mitigated, as explained in Section \ref{zeroUpd}. After the application of zero-type updates, non-holonomic constraints further enhance the solution. However, if the rover performs excessive turning several times as in the Pathfinder rover scenarios, non-holonomic updates may not be effective as explained in Section \ref{nonholo}. Note that, in this formulation, backward smoothing is shown to be most effective only if non-holonomic constraints are not applied mainly for the Husky rover. As such, this could be useful for a holonomic rover.

\begin{figure}[thpb]
      \centering
      \includegraphics[width=\columnwidth]{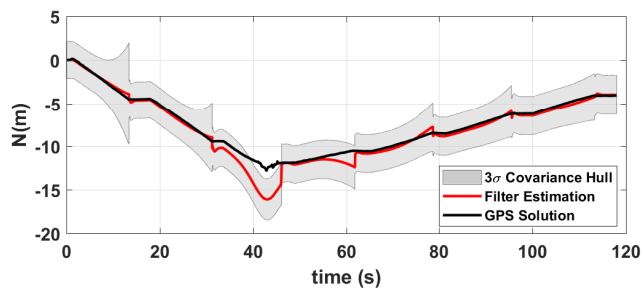}
      \caption{North error estimate without odometry aiding for Concrete-Turn case. Black line is GPS solution, red line is north position estimate, and gray region is $3\sigma$ covariance hull. Filter estimation exceeds the covariance bounds while turning with wheel slips. However, zero-type update reduces the cubic error growth to linear for positioning and correct most of the position drift when the rover stops as explained in Section \ref{zeroUpd}.}
      \label{northings2}
  \end{figure}

To better visualize the effects of the zero-type updates, the estimation of the north error for Concrete-Turn case is given in Fig.~\ref{northings2}. As shown in this figure, when the rover stops, the zero-type updates quickly reduce the growth of the position error and also correct most of the position drift.  

    \begin{figure}[b!]
      \centering
      \includegraphics[width=\columnwidth]{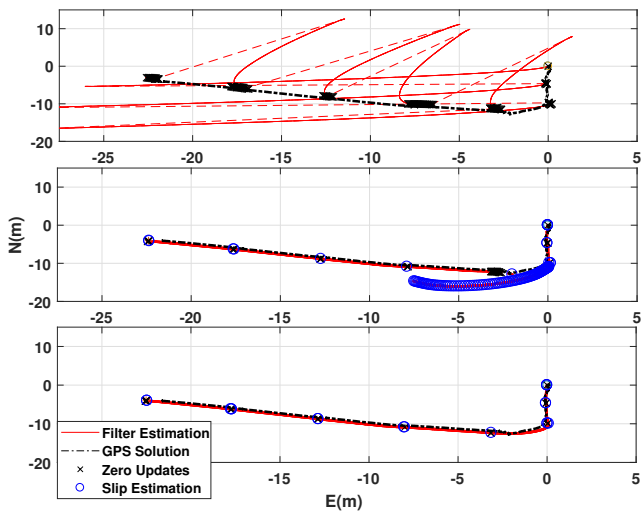}
      \caption{Ground track view of the ``Concrete-Turn" scenario. The top subplot shows the position estimation when only the zero-type update is applied which corrects the drifted position when the rover stops. The middle subplot demonstrates the position result when non-holonomic constraint updates are used along with zero-type updates, which corrects the lateral drifts while driving, but gives poor results while turning. The bottom figure is the result of the combination of the non-holonomic, zero-type, and wheel odometry solutions. The slip ratio, $i$ is checked to specify if the slip is significant (i.e., $|i|>0.3$) along with the Mahalanobis distance value at the same time-step. The heading rate is corrected during turning motion by fusing the wheel odometry measurements with INS solution along with the non-holonomic update.}
      \label{scene2}
  \end{figure}
  
To test the capabilities of the algorithm while turning, the Husky rover was exposed to a high rate of heading change after a forward drive on a flat concrete surface. Moreover, the rover was purposefully exposed to slippage by commanding an instantaneous maximum velocity (0.4 m/s) and braking after and before zero-type updates. This is evident when velocity anomaly detection and zero-type updates overlap, as shown in Fig.~\ref{scene2}. 
\begin{figure}[thpb]
  \centering
  \includegraphics[width=\columnwidth]{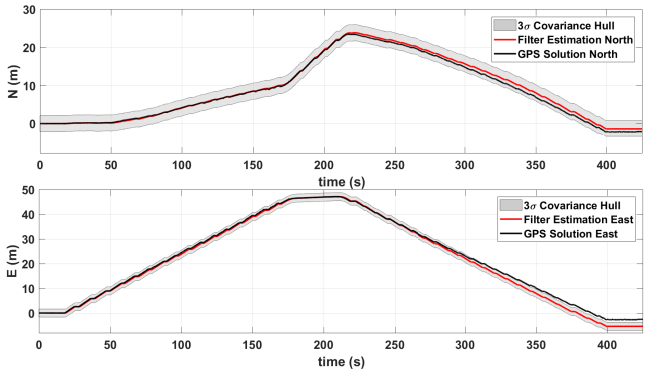}

  \caption{North (top) and east (bottom) error position estimates for Rough-Terrain case.  Black line is the GPS solution, red line is the position estimate, and the gray region is the $3\sigma$ covariance hull. Position is estimated by using all of the proposed measurement updates.}
  \label{covlong}
  \end{figure}
  \begin{figure}[thbp]
  \centering
  \includegraphics[width=\columnwidth]{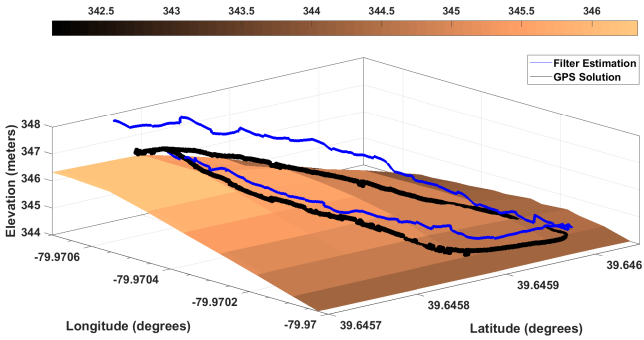}
  \caption{3D view of the rover traversal for Rough-Terrain in Latitude, Longitude, Height (LLH) coordinates. The GPS solution and filter estimation are plotted on USGS Digital Elevation Model (DEM) data of the traversed region. The 3D position error at the end amounts to less than 1.5\% of the traveled distance. }
  \label{figurelabelx}
  \end{figure}
        \begin{figure}[b!]
      \centering
      \includegraphics[width=\columnwidth]{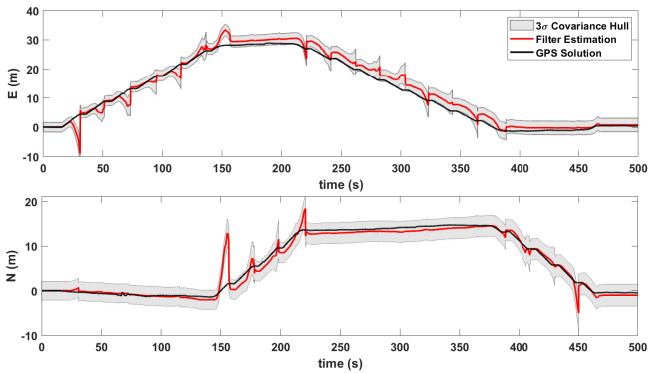}
      \caption{North and East error estimates I+Z+N+B scenario for Fast-Rectangle case. Black line is GPS estimated solution, red line is position estimate, and gray region is $3\sigma$ covariance hull. The highest peak values are the maximum worst case distance error. Zero-type updates recover the position solution when the rover stops.}
      \label{northeastFast}
  \end{figure}
    \begin{figure}[thbp]
  \centering
  \includegraphics[width=\columnwidth]{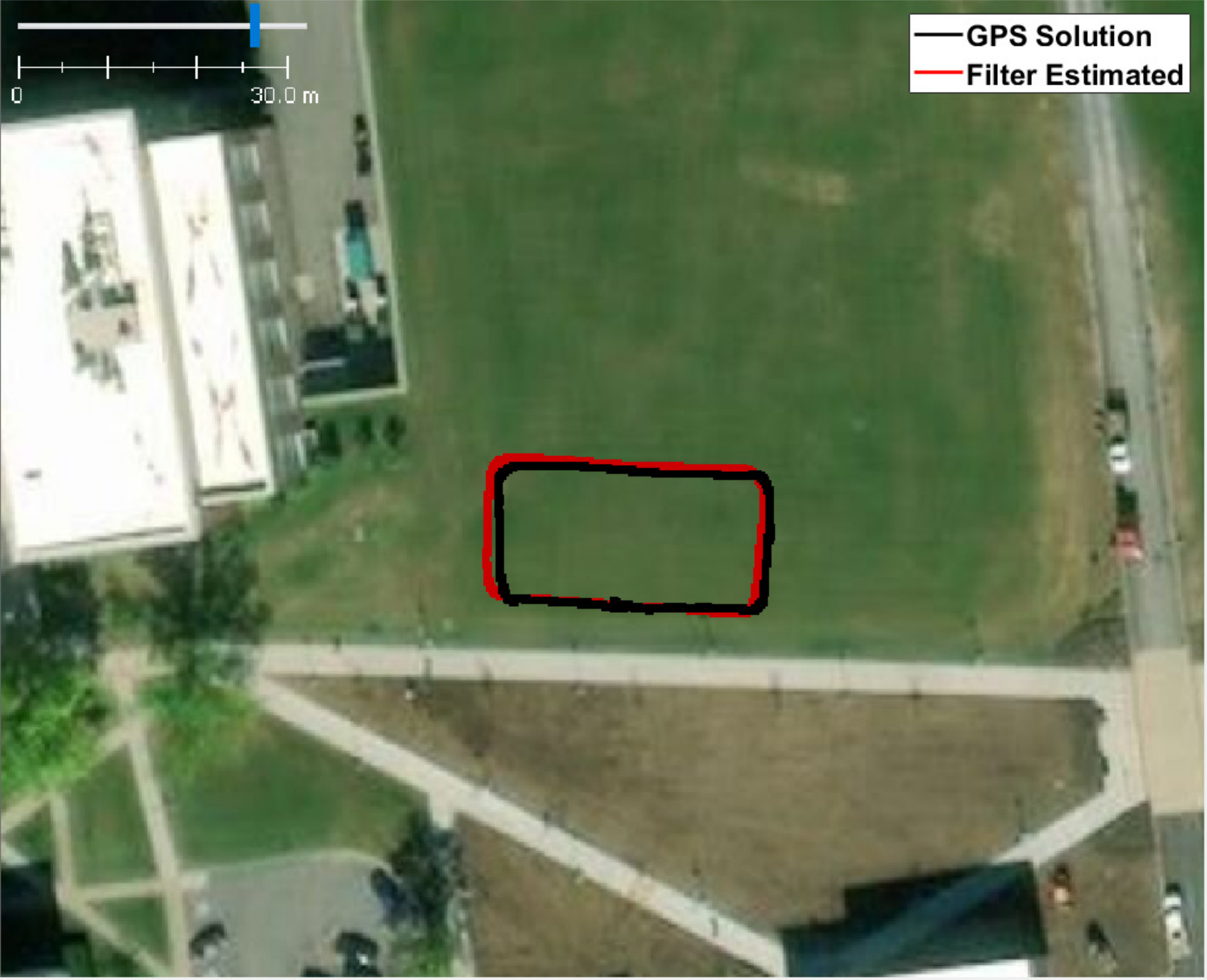}
  \caption{Ground track view of Pathfinder (Slow-Rectangle) scenario. Red line shows the filter estimated solution, black line shows the GPS solution. Overall 2D position error is 0.92 m with 0.84 m standard deviation.}
  \label{figurelabelMapPF}
  \end{figure}
  
  Although non-holonomic constraints handle most of the lateral motion drift, when the rover heading rate is significant (e.g., it exceeds 0.1 rad/s), lateral velocity constraint measurement updates become less reliable. However, when using wheel-odometry estimated heading rate measurements along with non-holonomic constraints to update the INS solution, less drift in position estimates is noticeable both when driving and when turning. This improvement is illustrated in Fig.~\ref{scene2}. Since the terrain of the ``Concrete-Turn" scenario is mostly easy to traverse and less prone to vibration, we also performed another test run on a harsh, non-flat, and diverse terrain. In the ``Rough-Terrain" case, the rover traversed a longer path. The estimated east and north position estimation with 3$\sigma$ error covariances are given in Fig.~\ref{covlong}. A 3D view of the positioning estimation when using all of the proposed updates is given in Fig.~\ref{figurelabelx}. For the sake of avoiding over-tuning the algorithm for a specific rover, the proposed algorithm is tested on a different testbed platform. With keeping the algorithm same for each rover, only the rover specific values are changed (e.g., wheel radius, IMU mount distance, wheelbase).
  
Furthermore, the algorithm is tested for higher velocity (0.8 m/s) as ``Fast-Rectangle" and lower velocity (0.2 m/s) as ``Slow-Rectangle" than the Husky rover's scenario velocities (0.4 m/s). One of the cases of the ``Fast-Rectangle" scenario, when there is no wheel odometry used, is given in Fig.~\ref{northeastFast} with north and east position estimations. A ground track view of the ``Slow-Rectangle" scenario when using all of the proposed updates is given in Fig.~\ref{figurelabelMapPF}.

\section{CONCLUSIONS}
\label{conclusions}

This work presents several approaches for enhancing the accuracy of wheeled planetary rover navigation. When using all of the presented update strategies together, the proposed approach was demonstrated to significantly reduce the rate of the rover navigation error growth. Besides, we showed that slippage could be detected by using a threshold on the velocity residuals and the slip ratio estimate. The primary value of this approach is that it can be used within current and future planetary rovers, as well as many other wheeled robots that frequently stops, to improve onboard localization performance without any hardware changes or major alterations in operations. 

The software developed for this paper, which is designed for use under ROS, and the datasets used for experimental validation have been made publicly available at: \\ \url{https://github.com/wvu-navLab/CLN}.  

Future developments of this work will include the extension of the proposed algorithm to autonomously initiating the navigation stops. Additionally, we are investigating ways to improve the backward smoothing for reprocessing the estimated navigation solution to mitigate the error that comes from slipped wheel odometry updates between zero-type updates and methods to improve the navigation solution by using onboard terrain classification techniques. 

\addtolength{\textheight}{-2cm}   

\section*{ACKNOWLEDGMENT}
The authors would like to acknowledge the support of NavLab and IRL members, especially Jonas A. Bredu, Chris Brindle, Shounak Das and Derek Ross at WVU. 
\bibliographystyle{IEEEtran}
\bibliography{core_nav.bbl}

\end{document}